%% file: main.tex
\documentclass[aps,pre,superscriptaddress,floatfix,tightenlines,showpacs,twocolumn,10pt,notitlepage]{revtex4-2}
\usepackage{amsmath}
\usepackage{amssymb}
\usepackage{amsthm}
\usepackage{bm}
\usepackage[shortlabels]{enumitem}
\usepackage{tabularx}
\usepackage{graphicx}
\usepackage{tikz-cd}
\usepackage{xcolor}
\usepackage{hyperref}
\usepackage[normalem]{ulem}
\usepackage[textsize=scriptsize,backgroundcolor=white]{todonotes}
\usepackage{import}

\hypersetup{
	colorlinks=true,
	linkcolor=blue,
	citecolor=blue,			
	urlcolor=blue,			
	filecolor=blue,
	anchorcolor = green,
}

\newcommand{\mb}[1]{\mathbf{#1}}
\newcommand{\mc}[1]{\mathcal{#1}}

\begin{document}
\title{Asymmetrically connected reservoir networks learn better}

\author{Shailendra K. Rathor}
\affiliation{Institute of Thermodynamics and Fluid Mechanics, Technische Universit\"at
Ilmenau, P.O.Box 100565, D-98684 Ilmenau, Germany}

\author{Martin Ziegler}
\affiliation{Micro- and Nanoelectronic Systems, Technische Universit\"at Ilmenau, P.O.Box 100565, D-98684 Ilmenau, Germany}

\author{Jörg Schumacher}
\affiliation{Institute of Thermodynamics and Fluid Mechanics, Technische Universit\"at
Ilmenau, P.O.Box 100565, D-98684 Ilmenau, Germany}
\affiliation{Tandon School of Engineering, New York University, New York City, NY 11201, USA}

\date{\today}

\begin{abstract}
We show that connectivity within the high-dimensional recurrent layer of a reservoir network is crucial for its performance. To this end, we systematically investigate the impact of network connectivity on its performance, i.e., we examine the symmetry and structure of the reservoir in relation to its computational power. Reservoirs with random and asymmetric connections are found to perform better for an exemplary Mackey-Glass time series than all structured reservoirs, including biologically inspired connectivities, such as small-world topologies. This result is quantified by the information processing capacity of the different network topologies which becomes highest for asymmetric and randomly connected networks.
\end{abstract}

\maketitle

Connectivity in artificial neural networks (ANNs) determines their performance, which is why many attempts are made to tailor them to the task at hand. The hand-picked and task-tailored connectivity of these networks makes ANNs extremely powerful, whereby such networks can outperform human performance in pattern recognition \cite{silver2016}, but with the disadvantage of a very high degree of specialisation, which makes learning new tasks difficult \cite{kudithipudi2022}. In addition, training ANNs is extremely energy-intensive and requires a large amount of well-labelled input data. Biological neural networks (BNNs) have a clear advantage here. The learning process is decentralised and self-organised, with the network structure constantly adapting to external influences. The price for this is the size and complexity of BNNs. In addition, the development and implementation of decentralised (local) learning and information processing mechanisms that specifically adapt the connectivity of the networks is extremely computationally intensive \cite{damicelli2022}.

In this context, the concept of reservoir computing (RC) is very exciting as it combines the advantages of ANNs and BNNs \cite{Jaeger2004}. In RC networks, data is embedded into a high-dimensional recurrent network (reservoir), which has random and sparse connectivity that does not need to be trained. Using a linear regression, only the connections between the read-out layer of the reservoir and the output layer are trained. This reduces training cost drastically, as only a small number of connections must be trained. As shown in the last decades the high dimensionality of the reservoir allows it to solve complex tasks~\cite{Pathak2017,Vlachas2020,Pandey2020,Heyder2022,Racca2023}.

However, a general link between network topology, modularity, or connectivity \cite{Newman2006, Carroll2019, Storm2022,Sun2022,viehweg_parameterizing_2023,Jaurigue2024,Jaurigue2024a,Micheli2024} and computing performance of recurrent networks while reaching the energy efficiency of BNNs is still not well understood \cite{Bassett2016,Bassett2017}. Canonically constructed networks with a reference to networks of our daily life, such as those by Watts and Strogatz \cite{Watts1998}, imply a symmetric connection matrix. Recently, several works~\cite{Kawai2017, Kawai2019, Kitayama2022} demonstrated that reservoirs constructed by the Watts-Strogatz (WS) method perform best when the rewiring probability is $p\sim 0.1$, i.e., for a small-world (SW) topology. Other studies, however, reported a relative independence of the performance with respect to the network structure~\cite{Fakhar2022}. Building on the SW architecture of the reservoir, more complex architectures such as multiple reservoirs or scale-free network topologies have been also explored~\cite{Deng2007}. Surprisingly little is, however, known about the impact of the symmetry of the reservoir matrix on the performance which is the central motivation of the present work.

In this work, we systematically investigate the impact of connectivity on the performance of RC networks, i.e., we analyse the reservoir performance in terms of symmetry and structure. We find that a completely random reservoir with asymmetric connections performs much better than all other symmetric reservoirs, including small-world topologies. For this, we investigated the performance of different reservoir topologies using the nonlinear time-delayed Mackey-Glass differential equation as a benchmark example. Our results are valid for open and closed loop regimes of the RC method \cite{Lukosevicius2023,Pfeffer2022}. We show that this behavior is reflected in the information processing capacity of the networks: a random reservoir with asymmetric connections has a higher information processing capacity than a symmetric reservoirs and reservoirs constructed by the WS method, even with a rewiring probability of 1. This shows that we must allow for a reservoir connectivity with the highest degree of complexity to ensure a high information processing capacity to achieve the maximum computing power.

\begin{figure}[t]
	\includegraphics[scale=0.95]{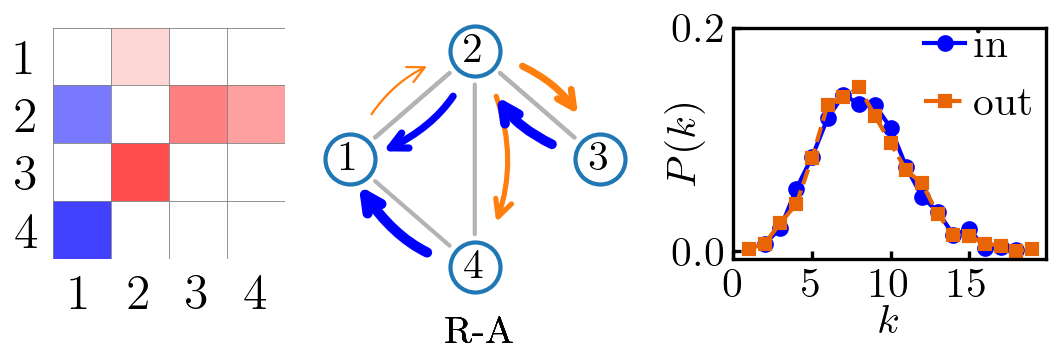}\\
	\includegraphics[scale=0.95]{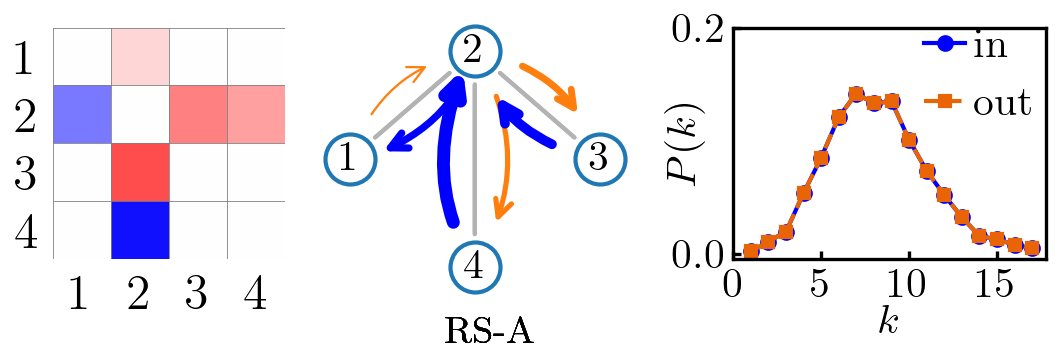}\\
	\includegraphics[scale=0.95]{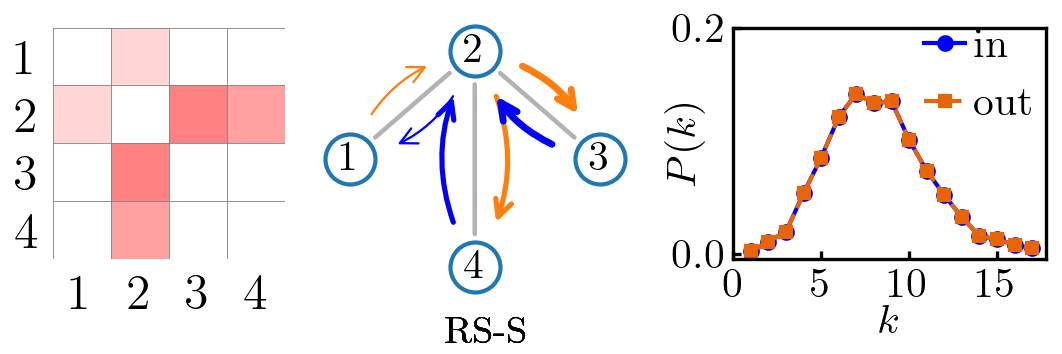}\\
	\includegraphics[scale=0.95]{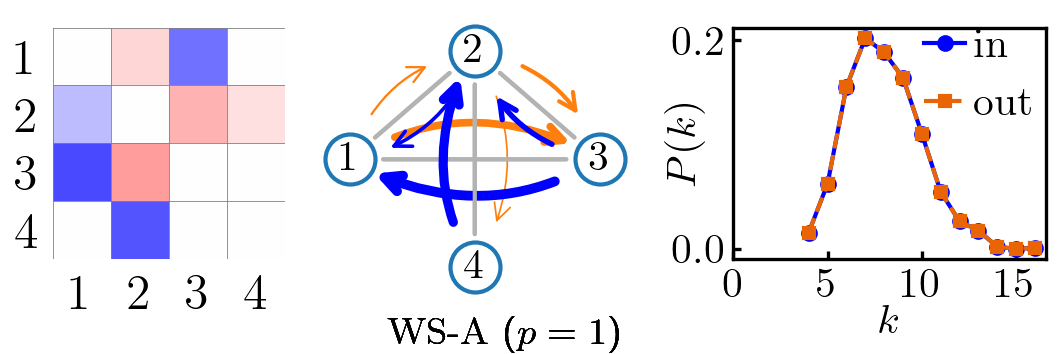}\\
	\hspace{-2mm}
	\includegraphics[scale=0.96]{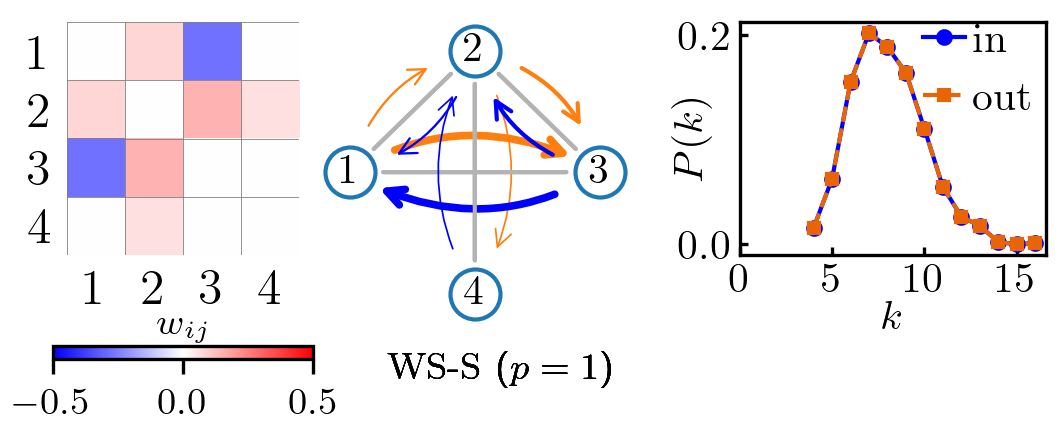}\\
	\vspace{-3mm}
	\caption{Representative configurations in a simple network with four nodes. The first column shows the reservoir matrix with color-coded weights. The second column shows the network graphs with color-coded weights. The thicker the arrow the higher weight magnitude. The third column displays the node degree distributions $P(k)$ with $N_r=1024$ nodes plotting incoming degrees (in) and outgoing degrees (out). Rows 1 to 5 from top to bottom show random--asymmetric (R-A), random symmetric--asymmetric (RS-A), random symmetric--symmetric (RS-S),  Watts-Strogatz--asymmetric (WS-A) with $p=1$, and Watts-Strogatz--symmetric (WS-S) with $p=1$. The first part of the acronym stands for the connectivity matrix $A$, the second one for the weight matrix $W^c$, see also eq. \eqref{eq:res}. Color bar holds for all cases.}
	\label{fig:network_configurations}
\end{figure}
{\em Reservoir network topologies.} Figure \ref{fig:network_configurations} gives an overview of the 5 investigated reservoir configurations for a small example network with 4 neurons. They are designed such that the connections {\em and} their strength (weights) can be modified separately. To this end, we consider the quadratic reservoir matrix $W$ as a Hadamard product (with element-wise multiplication) of a connection matrix $A$ and a weight matrix $W^c$,
\begin{equation}
W= A \odot W^c\,.
\label{eq:res}
\end{equation}
The fixed network connections, $i\to j$ and $j\to i$, are then encoded in the connection matrix $A$ and their weights, $w_{ij}$ and $w_{ji}$, in the weight matrix $W^c$ with neuron indices $i$ and $j$ ($i,j=1,..4$ in Fig. \ref{fig:network_configurations}); $w_{ij} \in \mathbb{R}$ chosen from a uniform distribution, $w_{ij}\in [-0.5, 0.5]$. The reservoir matrix $ W $ is subsequently normalized to set its spectral radius $\rho(W) = \rho_{\rm opt}$. The shown topologies follow from top to bottom, the most general case is a random network (R-A) with $A$ and $W^c$ both asymmetric, which even implies uni-directional connections, $w_{ij}\ne 0$, but $w_{ji}=0$. Secondly, a symmetrized connection matrix $A$ in combination with either an asymmetric or a symmetric weight matrix (RS-A and RS-S). Symmetric and asymmetric connectivity leads to undirected or directed networks \cite{Rubinov2010}. Finally, a (random) WS network with rewiring probability $p = 1$ which has by definition a symmetric $A$, but can obey either an asymmetric or a symmetric $W^c$ (WS-A and WS-S). We will also consider SW cases with $p<1$ in the WS cases.

The first column of Fig. \ref{fig:network_configurations} displays the entries of $W$ in our example color-coded by weight amplitude. The second column shows the network graphs, where in- and outgoing connections have orange and blue color. Arrow thickness is proportional to weight magnitude. The third column of Fig. \ref{fig:network_configurations} shows a further central property of all 5 reservoirs, the distribution of the node degree $k$ in the neuron network, $P(k)$, which we split into in- and outgoing. This distribution has been obtained from networks with $N_r=1024$ neurons, which we will apply in our analysis. For cases RS-A, RS-S, WS-A, and WS-S, in-and outgoing $P(k)$ have to coincide by construction.  Note that RS-A is similar to WS-A. Also R-A networks have a small mismatch between node degree distributions for in- and outgoing distributions. By construction, the support of $P(k)$ in the WS is smaller.

{\em Reservoir computing algorithm and dynamical system.} Following the RC computing scheme \cite{Jaeger2001}, the input data $\mb{u}_m$ enter the reservoir via a fixed random matrix $W^{\rm in}$. The dynamics of the reservoir neurons $\mb{r}_m$ is given by
\begin{equation}
\mb{r}_{m+1} = (1-\varepsilon)\mb{r}_{m} + \varepsilon\tanh \left(W  \mb{r}_{m} + W^{\rm in}  \mb{u}_{m+1}\right)\,.
\label{eq:RC}
\end{equation}
Here, index $m$ stands for a discrete equidistant time instant $t=m\Delta t=t_m$. The components of the random matrix $W^{\rm in}$ vary between $[-0.5,0.5]$. After the first $n_0$ steps the dynamics reaches the echo state; the reservoir becomes independent of the initial state~\cite{Jaeger2001,Yildiz2012}. We find that the optimal number for all reservoir types is $n_0 \sim 500$. The subsequent $n_{1}=2000$ outputs are collected for training the RC output layer. Therefore, the output layer weight matrix $W^{\rm out}$ is composed from the predicted output by $\mb{u}^p_{m+1}=W^{\rm out}\mb{r}_{m+1}$. The optimal output weights $W^{\ast {\rm out}}$ follow from the ridge regression scheme with a Tikhonov regularization parameter of $\gamma \sim {O}(10^{-9})$. Further hyperparameters are the reservoir size of $N_r=1024$ neurons, a leaking rate of $\varepsilon=0.7$, and spectral radius $\rho_{\rm opt}=1.25$. All networks are sparse with small fixed density of $0.008$ of active components in $W$. After training the network is run for a fixed time interval of $n_2 = 2000$, which defines the prediction phase.

To investigate the effects of the different reservoir structures in terms of computational performance, we will predict a one-dimensional time series $\mb{u}=u(t)$ which follows from the nonlinear time-delayed Mackey-Glass equation \cite{Mackey_1977}. The equation is given by
\begin{equation}
		\frac{du}{dt} = a \frac{u(t-\tau)}{1+u(t-\tau)^q} - b u(t)\,.
		\label{eq:MG17}
\end{equation}
Here are $a = 0.2$, $b = 0.1$, and $q = 10$ constants, while $\tau=17$ the delay time~\cite{SI}. The time-delayed equation \eqref{eq:MG17} describes a trajectory in an infinite-dimensional state space. For the given parameters, a chaotic attractor with a box counting dimension $2<D<3$ is formed \cite{Ziessler2019}. See Supplementary Information A for further time series generation details.

\begin{figure}[t]
		\includegraphics[scale=0.85]{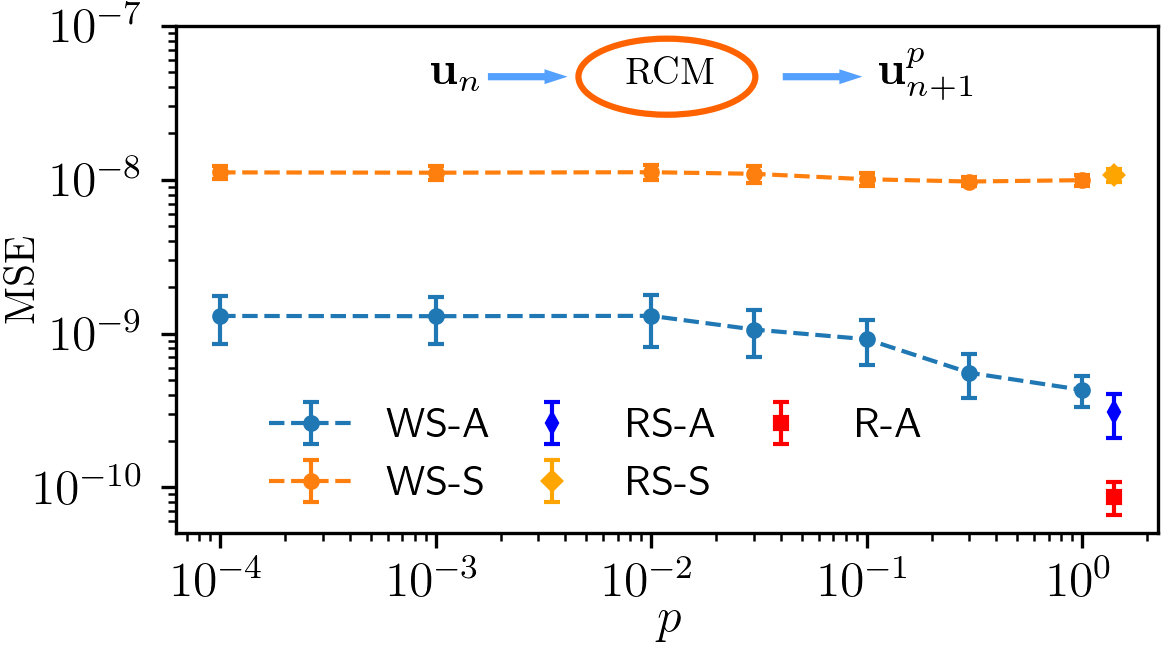}
		\caption[network comparison]{Open-loop reservoir performance as sketched in the central upper region. Double logarithmic plot of the median of the Mean-Square-Error (MSE) for all 5 reservoirs, for WS-A, and WS-S cases also dependent on rewiring probability $p$.  The error bars stand for the median absolute deviation (MAD), which is computed over an ensemble of $100$ differently initialized reservoir networks for each of the 5 configurations.}
		\label{fig:open-loop}
\end{figure}

{\em Open and closed loop performance.} We use the mean-square error (MSE) as a characteristic measure for the RC network performance. The MSE compares the ground truth (GT) $\hat{u}_m$ with model prediction $u^p_m$ and is given by ${\rm MSE} =  (1/M)\sum_{m=1}^{M}(\hat{u}_m - u^p_m)^2$. We start with the {\em open loop mode}, a one-step prediction (i.e., $\hat{u}_m = u_{m+1}$) of the reservoir, which is sketched in the top of Fig.~\ref{fig:open-loop}. There, we also plot the MSE (with $M=n_2$) as a function of the rewiring probability $p$ for both WS cases. It is seen that the best performance, i.e., the minimal MSE for WS networks, is obtained for $p=1$. This differs from \cite{Kitayama2022}. Errors, which are comparable but not equal to those of WS-S  and WS-A, follow for RS-S and RS-A, respectively. A first essential result of the analysis is that all networks with an asymmetric weight matrix perform significantly better than the corresponding symmetric ones. This is manifest by an MSE that is reduced by an order of magnitude. The MSE is reduced even by a further order of magnitude if we also take an asymmetric connection matrix $A$, being the case in R-A. Asymmetric random network configurations seem to perform best.

\begin{figure}[t]
	\includegraphics[scale=0.85]{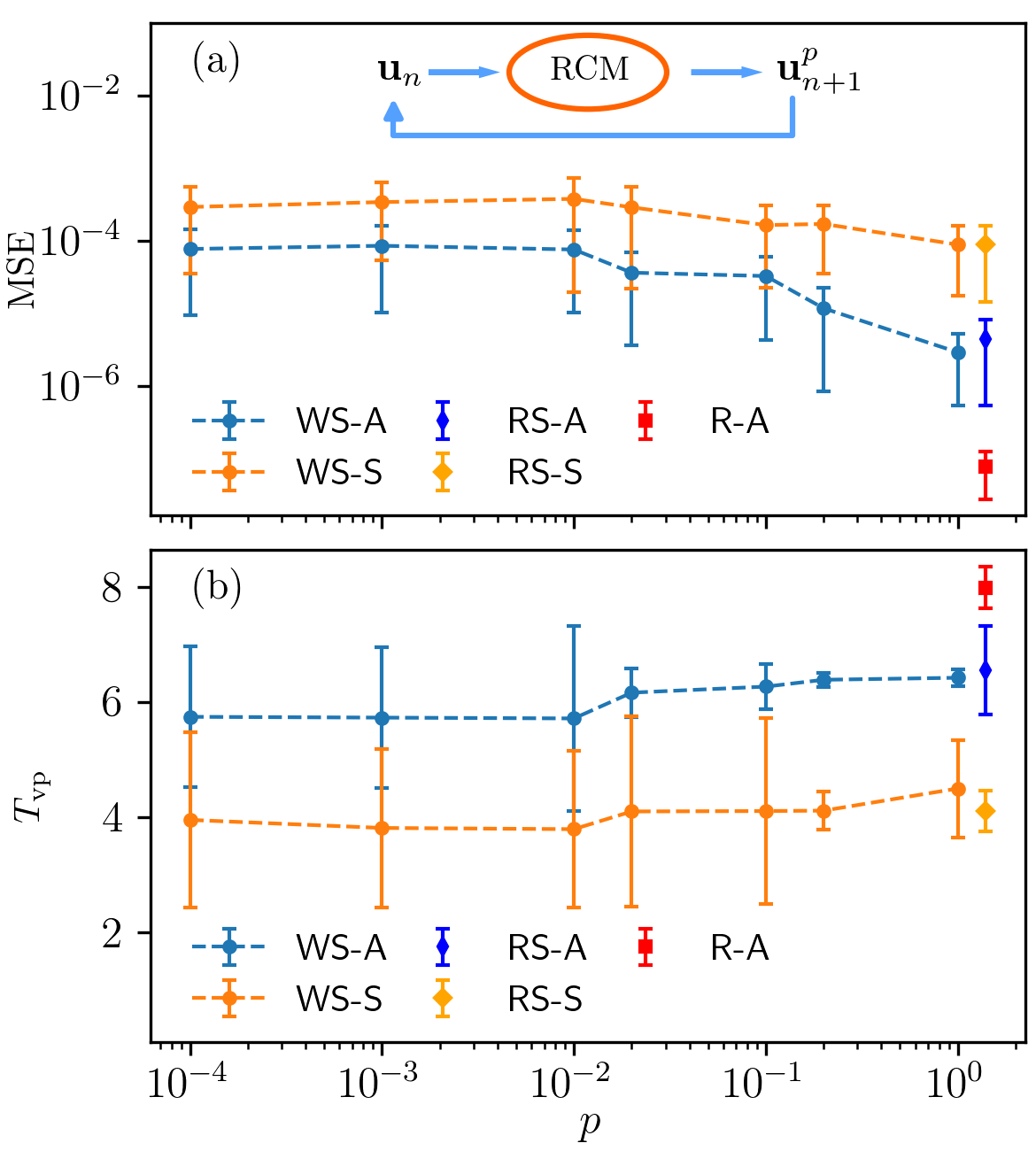}
	\caption[network comparison]{Closed-loop reservoir performance as sketched in the central upper region of the top panel. Top: Double logarithmic plot of the median of the Mean-Square-Error (MSE) from all network configurations similar to Fig. \ref{fig:open-loop}. Bottom: Median of the valid prediction time $T_{\rm vp}$ in units of Lyapunov time $1/\lambda_1$ for all 5 network configurations, for WS-A, and WS-S additionally as a function of $p$. The error bars stand now for the median absolute deviation (MAD), which is computed over an ensemble of $100$ differently initialized reservoirs for each specific network configuration.}
	\label{fig:close-loop}
\end{figure}

This result is basically confirmed for reservoirs which operate in the {\em close-loop mode}, as shown in Fig.~\ref{fig:close-loop}. The top panel of the figure replots MSE (with $M = n_2/4$) results. Despite the fact, that the MSE is always larger in the closed loop mode, where the RC output is re-fed for the next iteration, networks with symmetric $W^c$ perform again less well as those with asymmetric weights, even though the discrepancy is smaller. And again, R-A performs best. The predictive capability of a reservoir in closed loop mode is better characterised by the {\em valid prediction time} $T_{\rm vp}$ than by an MSE. To calculate  $T_{\rm vp}$ for the different reservoir types, one takes the normalized MSE, ${\rm NMSE}(m) = (\hat{u}_m - u^p_m)^2 / \sigma^2(\hat{u})$, where $\sigma^2(\hat{u})$ is the variance of the time series $ \hat{u} $ of GT. This leads to
\begin{equation}
	T_{\rm vp} = \max\{\lambda_1 t_m \,|\, {\rm NMSE}(\lambda_1 t_m) < {\rm NMSE}_{th}\},
\end{equation}
where $\lambda_1=0.007$ is the maximal Lyapunov exponent of the Mackey--Glass dynamics~\cite{Farmer1982}. The threshold is set to ${\rm NMSE}_{\rm th} = 0.25$. Times $T_{\rm vp}$ of the different reservoir types are shown in the bottom panel of Fig.~\ref{fig:close-loop}. All results are confirmed by this alternative performance measure; now reservoir R-A has the longest prediction horizon.

{\em Information processing capacity analysis.} To understand why random reservoirs with asymmetric connectivity perform better than symmetric ones, we calculated the task-independent information processing capacities of the different reservoir configurations. To this end, the {\em capacity} of a reservoir $\mc{R}$ driven by one-dimensional external input $ \xi $ is defined as the capability to reconstruct a time series of $M$ data points, $\{\hat{u}_m\}^M_{m=1}$, by the reservoir $\mc{R}$ generating an output series $\{u_m^p\}^M_{m=1}$ and given by, see~\cite{Dambre2012,Huelser2023},
\begin{equation}
	C_M[\mc{R}, \hat{\bm u}] = 1 - \frac{\sum_{m=1}^M (\hat{u}_m-u^p_m)^2}{ \sum_{m=1}^M\hat{u}^2_m}\,.
 \label{eq:cap}
\end{equation}
If the time series $\{u_m^p\}^M_{m=1}$ is a perfect reconstruction of $\{\hat{u}_m\}^M_{m=1}$, then $C_M[\mc{R},
\hat{\bm u}] = 1$; if the reservoir fails to reconstruct $\hat{u}_m$ then $C_M[\mc{R},\hat{\bm u}] =
0$. The capacity \eqref{eq:cap} still depends on the input $\xi$. Independence of a particular input
is obtained by taking a random input that is sampled from a uniform distribution \cite{Jaeger2001,Dambre2012}. Since the
reconstruction is done by a linear estimator, $u^p_m = W^{\ast{\rm out}} \mb{r}_m$, all the nonlinear processing is then
performed by the reservoir, and an appropriately defined measure, the {\em total information processing capacity} (IPC),
becomes an  exclusive property of the specific reservoir $\mc{R}$, see also ref. \cite{Huelser2023}.

In the Supplementary Information B \cite{SI}, we explain the technical details which are required to define the
task-independent IPC. In a nutshell, we build a basis of the Hilbert space ${\cal H}$ of fading memory functions by
taking product of Legendre polynomials of past input. Therefore, the $ n $-th basis $ {\varphi_n} (\xi_m^{-\infty}) $
is a function of a sequence of past inputs that reach from an instant at time $m$ into the past, i.e., $\xi_m^{-\infty} = \xi_m,
\xi_{m-1}, \dots,\xi_{m-j}, \dots, \xi_{m-\infty}$. The total IPC is an infinite sum of IPCs of processing basis functions $ \varphi_n $ that are polynomials of different degrees $d$. For the practical analysis, neither the time series can go
back to an instant $m-\infty$ nor the degree $d$ of the  test polynomials can be increased infinitely. Thus the
following approximation for a total IPC results, see again \cite{SI}
\begin{equation} \mbox{IPC}_{\rm
	total}=\sum_{d=1}^{\infty} \mbox{IPC}_{d} \approx \sum_{d=1}^D \sum_{n=1}^{\mc N_{d}} C_M[{\mathcal R}, {\varphi}_n(\xi_m^{-J})]\,. \label{eq:IPC}
\end{equation}
Here, $D$ is a cutoff in degree, the delay $m-\infty$ is substituted by $m-J$, and $ \mc N_{d} $ is the reduced number of basis functions of degree $ d $. It was shown that the IPC$_{\rm total} \le N_r$ is bounded from above by the number of nodes of the reservoir $N_r$, which allows us to compare the (approximated) total  IPC of different reservoirs with the same set of input functions, even though the appropriateness of this measure was critically reflected in ref. \cite{Huelser2023} for time-multiplexed reservoirs. Thus IPC$_{\rm total}$ gives us a measure of the internal reservoir architecture which is required to rationalize the results from Figs. \ref{fig:open-loop} and \ref{fig:close-loop}.

To compute IPCs, we take combination $(d, J)$ in $\{(1, 2000), (2, 300), (3, 50), (4, 30), (5, 15)\}$ and input size $M = 9 \times 10^5$. Further, to avoid overestimation of the capacity due to finite $M$, we set the capacities less than a small threshold to exactly zero. In the Supplementary Information C, we explain the calculation of IPC$_d$ in detail for the case of $d=3$ \cite{SI}.

\begin{figure}
	\includegraphics[scale=0.95]{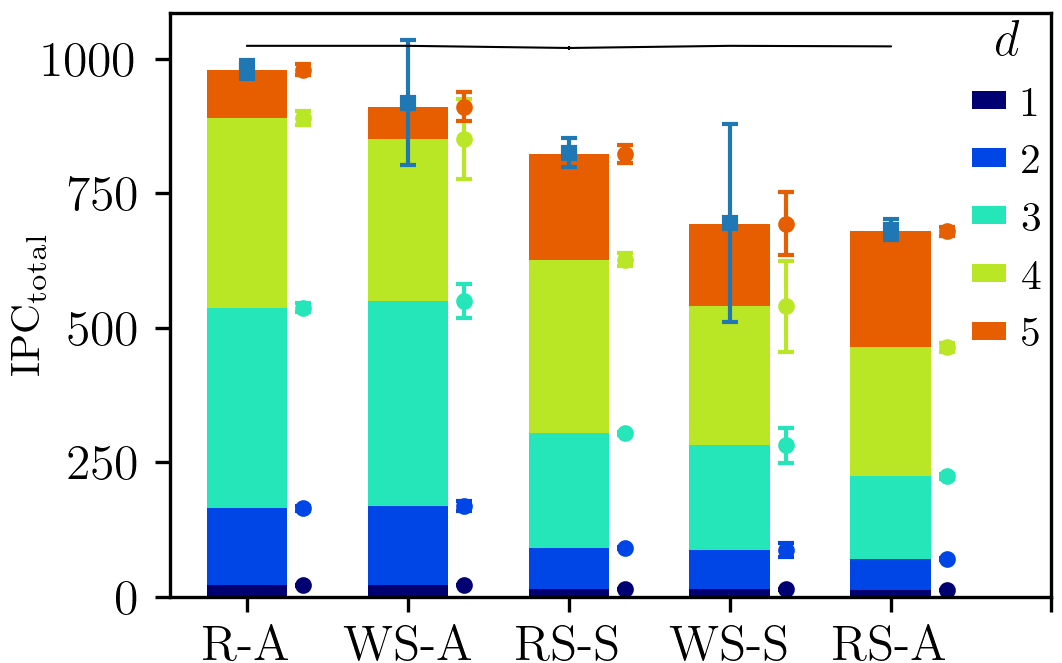}
	\caption{Information processing capacities of degree $d$ and accumulated total information processing capacity $\mbox{IPC}_{\rm total}$ for upper degree bound $D=5$. The median of total information processing capacities IPC$_{\rm total}$ (blue squares) and their decomposition with respect to the degree $d$ (see legend) for the 5 reservoir configurations. The error bars represent the median absolute deviation, which is computed over an ensemble of $40$ differently initialized networks for each specific network configuration. The solid line shows the upper bound of $ \rm IPC_{total}\le N_r =1024$ for the reservoir configurations.}
	\label{fig:ipc}
\end{figure}
Figure~\ref{fig:ipc} compares the statistics of $ \rm IPC_{total} $ of different reservoir configurations over $ 40 $ realizations of $ A $ and $ W^c $. The $ \rm IPC_{total} $ of asymmetrically connected reservoir $ \mathcal{R} =$ R-A  is larger than the capacity of any of the remaining 4 symmetrically connected reservoirs. The $ \rm IPC_{total} $ of $ \mathcal{R}= $ WS-A displays a relatively large error bar (that covers magnitudes as high as R-A) which implies that WS-A is highly sensitive to the initialization of reservoir. This is true for the symmetric WS network as well. Note that the markers with error bars next to each histogram column represent the median IPC to reconstruct a polynomial of respective degree. This sensitivity on the initialization of reservoir increases with the degree (and thus with the nonlinearity) as shown by exactly these error bars. WS networks are in general unstable with respect to asymmetric connection networks. Furthermore, note that the symmetric reservoirs $\mathcal{R} = \{$WS-S, RS-S, RS-A$\}$ require relatively more capacity to reconstruct highly nonlinear functions, e.g., polynomials of degree $5$.

{\em Conclusions.}
RC computing provides a biologically inspired way to reduce the training overhead of setting connectivity in an ANN using a static high-dimensional recurrent network (reservoir). In this letter, we have shown that reservoirs with random and asymmetric connections perform better than all structured reservoirs, including biologically inspired small-world topologies. This result was analysed for Mackey-Glass time series and quantified by the information processing capacity of the different networks investigated. This shows that the highest performance is achieved by maximising the disorder of the network structure, which is in line with biological networks. Future directions to extend our work comprises nonlinear models with more degrees of freedom and networks with binary connections \cite{Ma2023}.

{\em Acknowledgments.} The work of S.K.R. and J.S. is funded by the European Union (ERC, MesoComp, 101052786). Views and opinions expressed are however those of the authors only and do not necessarily reflect those of the European Union or the European Research Council. M.Z. acknowledges financial support by Deutsche Forschungsgemeinschaft (DFG, German Research Foundation) – Project RECOMMEND Project number 536063366 as part of the DFG priority program DFG-SPP 2262 MemrisTec – Project number 422738993.
We thank Lina Jaurigue for helpful discussions.

\bibliography{main}



\clearpage
\section{Supplementary material}
\subsection{Mackey--Glass time series generation}
The Mackay--Glass time series \cite{Mackey_1977} is generated by iterating the discrete approximation of the delay differential equation, see eq. \eqref{eq:MG17} in the main text, which is described by Grassberger and Proccacia~\cite{Grassberger1983}. The initial conditions are taken randomly from a uniform distribution and an initial transient of $250,000$ integration steps is discarded before training and test data are used. The time step of the Grassberger-Procaccia iteration is $ \delta t=1.7 \times 10^{-2} $. Further, the time series is downsampled to obtain a step size of $\Delta t=1$. We used rescaled time series with amplitudes  $u(t)\in [-1,1]$.

\subsection{Calculation of the total information processing capacity}
The outlined construction procedure of the total information processing capacity $\mbox{IPC}_{\rm total}$ follows closely the references \cite{Dambre2012, Huelser2023} and is adapted to the notation of the present manuscript. Starting point is the infinite-dimensional Hilbert space ${\cal H}$ of fading memory functions, which is spanned by a complete set of basis functions $\{\varphi_n\}_{n=1,2,\dots}$ with an index $n$ that is specified later. The arguments $\xi$ of the basis functions will be uniformly distributed random numbers $\xi$ sampled from the interval $[-1,1]$. More specifically, we denote input argument time series by
\begin{align}
\xi_1^{-\infty}&=(\xi_1, \xi_0, \xi_{-1}, \xi_{-2}, \dots, \xi_{1-\infty})\nonumber \\
\xi_2^{-\infty}&=(\xi_2, \xi_1, \xi_{0}, \xi_{-1}, \dots, \xi_{2-\infty})\,,
\label{eq:app2}
\end{align}
and thus generally
\begin{align}
\xi_k^{-\infty}&=(\xi_k, \xi_{k-1}, \xi_{k-2}, \xi_{k-3}, \dots, \xi_{k-\infty})\,.
\label{eq:app2a}
\end{align}
They contain the input $\xi_k$ and all corresponding past random inputs at time instance $k$. The basis functions are composed of products of Legendre polynomials $P_{\ell}(\xi)$ of degree $\ell$. The $n$-th basis function is given by
\begin{align}
\varphi_n(\xi_k^{-\infty}) &= \prod_{j=0}^{\infty} P_{d_{n}^{j}} (\xi_{k-j}) \nonumber\\
&=P_{d_{n}^{1}}(\xi_{k-1}) P_{d_{n}^{2}}(\xi_{k-2}) P_{d_{n}^{3}}(\xi_{k-3}) \dots\,,
\label{eq:app3}
\end{align}
where the multi-index $n$ stands for $n=\{d_n^0,d_n^1,d_n^2,\dots\}$, i.e., it contains the degrees $d_n^{j}$ of the Legendre polynomials in the product. The total degree corresponds thus eventually to the degree of the nonlinearity, as we will see further below. Only a finite number of all $d_n^j$ in the multi-index $n$ is non-zero. Recall that a Legendre polynomial of degree $\ell$ is constructed by the Rodrigues' formula, which is given by
\begin{align}
P_{\ell}(\xi) = \frac{1}{2^{\ell} \ell\,!}\, \frac{d^{\ell}}{d\xi^{\ell}}\left[(\xi-1)^{\ell}\right]\,.
\label{eq:app4}
\end{align}
One defines the total degree of the basis vector of the Hilbert space, $\varphi_n\in {\cal H}$, by
\begin{align}
d(\varphi_n)= \sum_{j=0}^{\infty} d_n^{j} <\infty\,.
\label{eq:app5}
\end{align}
One can now calculate the information processing capacities (IPCs) of degree $d$, which can have linear, quadratic, cubic, or higher order degree, by
\begin{align}
\mbox{IPC}_{d}=\sum_{n =1}^{\infty} \left[1 - \frac{\sum_{m=1}^M (\varphi_n(\xi_m^{-\infty})-u^p_m)^2}{ \sum_{m=1}^M\varphi^2_n(\xi_m^{-\infty})} \right]\Bigg|_{d(\varphi_n)=d}\,.
\label{eq:app6}
\end{align}
See again eq. \eqref{eq:cap} in the main text for the definition of the capacity. Finally, the total information processing capacity is given by the sum of all IPCs of degree $d$
\begin{align}
\mbox{IPC}_{\rm total}=\sum_{d=1}^{\infty} \mbox{IPC}_{d}\,.
\label{eq:app7}
\end{align}
This gives the first expression on the right hand side of eq. \eqref{eq:IPC} in the main manuscript. For a practical evaluation, all infinite products and sums will become finite. In the present work, $\mbox{IPC}_{\rm total}=\mbox{IPC}_1+ \dots + \mbox{IPC}_5$ which saturates the upper bound already. This implies that we took $d\le D=5$, see again the last term in eq. \eqref{eq:IPC}. Therefore, the following lowest-degree Legendre polynomials are of interest,
\begin{eqnarray*}
    P_0(\xi) &=& 1,\\
    P_1(\xi) &=& \xi, \\
    P_2(\xi) &=&  \frac{1}{2}(3\xi^2 - 1), \\
    P_3(\xi) &=&  \frac{1}{2}(5\xi^3 - 3\xi), \\
    P_4(\xi) &=&  \frac{1}{8}(35\xi^4 - 30\xi^2 + 3), \\
    P_5(\xi) &=&  \frac{1}{8}(63\xi^5 - 70\xi^3 + 15\xi).
\end{eqnarray*}
In practice, $ P_{\alpha}, $ for $ \alpha \ge 2 $ with $ P_0=1$ and $P_1=\xi$, are computed by Bonnet's recursion formula:
\begin{equation}
    (\alpha + 1) P_{\alpha + 1}(\xi) = (2\alpha + 1) \xi P_\alpha(\xi) - \alpha P_{\alpha - 1}(\xi).
\end{equation}
These Legendre polynomials satisfy the orthonormalization condition:
\begin{equation}
    \int_{-1}^{1} P_{\alpha}(\xi) P_{\alpha^\prime}(\xi) d\xi = \delta_{\alpha, \alpha ^\prime}.
\end{equation}
\subsection{Detailed explanation for the case of IPC$_3$}
In the following, we illustrate the computation of IPC$_{d}$ in eq.~\eqref{eq:app6} with
$d = 3$ as an example.

\begin{enumerate}
    \item First, we have to determine all possible combinations
$ C_d $  of Legendre polynomial products, such that the total degree sums up to $d=3$. The set of
possible degrees for $d = 3$ is $\{(1,1,1), (1,2), (3)\}$, i.e., three terms in the product are linear polynomials or one is linear and one quadratic, or only one is cubic. In other words, the degree combination $ C_3 = (1,2)$ implies that, in eq.~\eqref{eq:app3}, there are $P_1$ and $ P_2 $ evaluated together at different delayed inputs and all other Legendre polynomials are $P_0=1$ (see Fig.~\ref{fig:ipcd32}).\\
    \item Secondly, we construct the basis functions $ \varphi_n(\xi_k^{-\infty})$ using the above
determined Legendre polynomials, which are fed with arguments $\xi_{k-j}$ for
delay indices $j = 0,1,2,\dots,J$, without repeating $j$ in $\xi_{k-j}$. \\
The total number ${\mc{N}_d}$ of possible products for degree $d$ in eq. \eqref{eq:app3} is thus given by
\begin{align}
{\mc{N}_d}=\sum_{q_i=1}^d {\cal M}_i \times \frac{(J+1)!}{(J+1-q_i)! q_i!}\,.
\end{align}
The first factor under the sum recognizes the multiplicities ${\cal M}_i$ for the selected subset of polynomial degrees due to permutations, the second term counts the combinations. In our example of $d=3$ and for $ J = 50 $ (see the main text), we get: the degree combination $C_3=(1,1,1) $, which has $ q_1=3$ elements and a multiplicity of ${\cal M}_1=1$, is associated with  $ 20825 $ basis functions. Similarly, the degree combination $ C_3 = (1,2)$ with $ q_2=2 $ and multiplicity ${\cal M}_2=2$ results in  $2550$ products. Finally, $ C_3 = (3)$ with $q_3 = 1 $ and multiplicity ${\cal M}_3=1$ adds $ 51 $ basic functions. The construction of  $ \varphi_n(\xi_k^{-50})$ is also illustrated in Figs.~\ref{fig:ipcd31} and ~\ref{fig:ipcd32}.
\begin{figure}[htpb]
    \import{./}{ipc-d3-1.tex}
    \caption{Different degree combinations (1,1,1) are illustrated. First row shows the sequence of delayed inputs $ \xi_k^{-50}$. Each subsequent row has a unique arrangement of Legendre polynomials $ P_1 $  with argument as the corresponding element from the first row. The number of arrangements is $\binom {51}{3} = 20825$. Here, 1 represents the first Legendre polynomial $ P_0=1$.}
    \label{fig:ipcd31}
\end{figure}
\begin{figure}[htpb]
    \import{./}{ipc-d3-2.tex}
    \caption{Different degree combinations (1,2) are illustrated. First row shows the sequence of delayed inputs $ \xi_k^{-50}$. Each subsequent row has a unique arrangement of Legendre polynomials $ P_1 $ and $ P_2 $  with argument as the corresponding element from the first row. Rows 2 and 3 show the multiplicity $ \mc{M}_2 = 2 $ for the given choice of arguments $\xi_k$ and $ \xi_{k-1} $ of polynomials of non-zero degree. The number of arrangements is $2.\binom {51}{2} = 2550$. Here, 1 represents the first Legendre polynomial $ P_0=1$.}
    \label{fig:ipcd32}
\end{figure}
\item Finally, each of the $ \mc{N}_d = 23426 $  basis functions is reconstructed by the RC model to evaluate IPC$_3$, which is the sum of all the capacities corresponding to the basis functions, refer to eq.~\eqref{eq:app6}. The capacity of processing a basis function is set to zero if it is smaller than a threshold value.
\end{enumerate}

\end{document}

%% file: ipc-d3-1.tex
\begin{tikzpicture}[scale=1, transform shape, delaybox/.style={rectangle, thick, draw=black!60, fill=black!5, text width=8mm, minimum width=8mm, minimum height=5mm, align=center}, box/.style={rectangle, thick, draw=blue!60, fill=blue!5, text width=8mm, minimum width=8mm, minimum height=5mm, align=center}]

    \node at (0,0) [delaybox] (box) {$ \xi_{k} $};
    \node at (1,0) [delaybox] (box) {$ \xi_{k-1} $};
    \node at (2,0) [delaybox] (box) {$ \xi_{k-2} $};
    \node at (3,0) [delaybox] (box) {$ \xi_{k-3} $};
    \node at (4,0) [delaybox] (box) {$ \xi_{k-4} $};
    \node at (5,0) [delaybox, draw=white, fill=white, text width=6mm] (box) {$ \cdots $};
    \node at (6,0) [delaybox] (box) {$ \xi_{k-49} $};
    \node at (7,0) [delaybox] (box) {$ \xi_{k-50} $};

    \node at (0,-0.7) [box] (box) {$ P_1 $};
    \node at (1,-0.7) [box] (box) {$ P_1$};
    \node at (2,-0.7) [box] (box) {$ P_1$};
    \node at (3,-0.7) [box] (box) {$ 1 $};
    \node at (4,-0.7) [box] (box) {$ 1 $};
    \node at (5,-0.7) [box, draw=white, fill=white, text width=6mm] (box) {$ \cdots $};
    \node at (6,-0.7) [box] (box) {$ 1$};
    \node at (7,-0.7) [box] (box) {$ 1$};

    \node at (0,-1.4) [box] (box) {$ 1 $};
    \node at (1,-1.4) [box] (box) {$ P_1 $};
    \node at (2,-1.4) [box] (box) {$ P_1 $};
    \node at (3,-1.4) [box] (box) {$ P_1 $};
    \node at (4,-1.4) [box] (box) {$ 1 $};
    \node at (5,-1.4) [box, draw=white, fill=white, text width=6mm] (box) {$ \cdots $};
    \node at (6,-1.4) [box] (box) {$ 1$};
    \node at (7,-1.4) [box] (box) {$ 1 $};

    \node at (2,-2.1) [box, draw=white, fill=white, minimum height=5mm] (box) {$ \vdots $};
    \node at (6.5,-2.1) [box, draw=white, fill=white, minimum height=5mm] (box) {$ \vdots $};

    \node at (0,-3.0) [box] (box) {$ P_1 $};
    \node at (1,-3.0) [box] (box) {$ 1 $};
    \node at (2,-3.0) [box] (box) {$ 1 $};
    \node at (3,-3.0) [box] (box) {$ P_1 $};
    \node at (4,-3.0) [box] (box) {$ 1 $};
    \node at (5,-3.0) [box, draw=white, fill=white, text width=6mm] (box) {$ \cdots $};
    \node at (6,-3.0) [box] (box) {$ P_1$};
    \node at (7,-3.0) [box] (box) {$ 1 $};

    \node at (2,-3.7) [box, draw=white, fill=white, minimum height=5mm] (box) {$ \vdots $};
    \node at (6.5,-3.7) [box, draw=white, fill=white, minimum height=5mm] (box) {$ \vdots $};

    \node at (0,-4.6) [box] (box) {$ 1 $};
    \node at (1,-4.6) [box] (box) {$ 1 $};
    \node at (2,-4.6) [box] (box) {$ 1 $};
    \node at (3,-4.6) [box] (box) {$ 1 $};
    \node at (4,-4.6) [box] (box) {$ P_1 $};
    \node at (5,-4.6) [box, draw=white, fill=white, text width=6mm] (box) {$ \cdots $};
    \node at (6,-4.6) [box] (box) {$ P_1$};
    \node at (7,-4.6) [box] (box) {$ P_1 $};
\end{tikzpicture}

%% file: ipc-d3-2.tex
\begin{tikzpicture}[scale=1, transform shape, delaybox/.style={rectangle, thick, draw=black!60, fill=black!5, text width=8mm, minimum width=8mm, minimum height=5mm, align=center}, box/.style={rectangle, thick, draw=blue!60, fill=blue!5, text width=8mm, minimum width=8mm, minimum height=5mm, align=center}]

    \node at (0,0) [delaybox] (box) {$ \xi_{k} $};
    \node at (1,0) [delaybox] (box) {$ \xi_{k-1} $};
    \node at (2,0) [delaybox] (box) {$ \xi_{k-2} $};
    \node at (3,0) [delaybox] (box) {$ \xi_{k-3} $};
    \node at (4,0) [delaybox] (box) {$ \xi_{k-4} $};
    \node at (5,0) [delaybox, draw=white, fill=white, text width=6mm] (box) {$ \cdots $};
    \node at (6,0) [delaybox] (box) {$ \xi_{k-49} $};
    \node at (7,0) [delaybox] (box) {$ \xi_{k-50} $};

    \node at (0,-0.7) [box] (box) {$ P_1 $};
    \node at (1,-0.7) [box] (box) {$ P_2$};
    \node at (2,-0.7) [box] (box) {$ 1$};
    \node at (3,-0.7) [box] (box) {$ 1 $};
    \node at (4,-0.7) [box] (box) {$ 1 $};
    \node at (5,-0.7) [box, draw=white, fill=white, text width=6mm] (box) {$ \cdots $};
    \node at (6,-0.7) [box] (box) {$ 1$};
    \node at (7,-0.7) [box] (box) {$ 1$};

    \node at (0,-1.4) [box] (box) {$ P_2 $};
    \node at (1,-1.4) [box] (box) {$ P_1$};
    \node at (2,-1.4) [box] (box) {$ 1$};
    \node at (3,-1.4) [box] (box) {$ 1 $};
    \node at (4,-1.4) [box] (box) {$ 1 $};
    \node at (5,-1.4) [box, draw=white, fill=white, text width=6mm] (box) {$ \cdots $};
    \node at (6,-1.4) [box] (box) {$ 1$};
    \node at (7,-1.4) [box] (box) {$ 1$};

    \node at (2,-2.1) [box, draw=white, fill=white, minimum height=5mm] (box) { $ \vdots $ };
    \node at (6.5,-2.1) [box, draw=white, fill=white, minimum height=5mm] (box) { $ \vdots $ };

    \node at (0,-3.0) [box] (box) {$ 1 $};
    \node at (1,-3.0) [box] (box) {$ P_1 $};
    \node at (2,-3.0) [box] (box) {$ 1 $};
    \node at (3,-3.0) [box] (box) {$ 1 $};
    \node at (4,-3.0) [box] (box) {$ 1 $};
    \node at (5,-3.0) [box, draw=white, fill=white, text width=6mm] (box) {$ \cdots $};
    \node at (6,-3.0) [box] (box) {$ P_2$};
    \node at (7,-3.0) [box] (box) {$ 1 $};

    \node at (0,-3.7) [box] (box) {$ 1 $};
    \node at (1,-3.7) [box] (box) {$ P_2 $};
    \node at (2,-3.7) [box] (box) {$ 1 $};
    \node at (3,-3.7) [box] (box) {$ 1 $};
    \node at (4,-3.7) [box] (box) {$ 1 $};
    \node at (5,-3.7) [box, draw=white, fill=white, text width=6mm] (box) {$ \cdots $};
    \node at (6,-3.7) [box] (box) {$ P_1$};
    \node at (7,-3.7) [box] (box) {$ 1 $};

    \node at (2,-4.4) [box, draw=white, fill=white, minimum height=5mm] (box) { $ \vdots $ };
    \node at (6.5,-4.4) [box, draw=white, fill=white, minimum height=5mm] (box) { $ \vdots $ };

    \node at (0,-5.3) [box] (box) {$ 1 $};
    \node at (1,-5.3) [box] (box) {$ 1 $};
    \node at (2,-5.3) [box] (box) {$ 1 $};
    \node at (3,-5.3) [box] (box) {$ 1 $};
    \node at (4,-5.3) [box] (box) {$ 1 $};
    \node at (5,-5.3) [box, draw=white, fill=white, text width=6mm] (box) {$ \cdots $};
    \node at (6,-5.3) [box] (box) {$ P_1$};
    \node at (7,-5.3) [box] (box) {$ P_2 $};

    \node at (0,-5.8) [box] (box) {$ 1 $};
    \node at (1,-5.8) [box] (box) {$ 1 $};
    \node at (2,-5.8) [box] (box) {$ 1 $};
    \node at (3,-5.8) [box] (box) {$ 1 $};
    \node at (4,-5.8) [box] (box) {$ 1 $};
    \node at (5,-5.8) [box, draw=white, fill=white, text width=6mm] (box) {$ \cdots $};
    \node at (6,-5.8) [box] (box) {$ P_2$};
    \node at (7,-5.8) [box] (box) {$ P_1 $};
\end{tikzpicture}